\pdfoutput=1

\documentclass[11pt]{article}

\usepackage{ACL2023}

\usepackage{times}
\usepackage{latexsym}

\usepackage{amsfonts}

\usepackage{graphicx}
\usepackage{subfigure}

\usepackage{siunitx}
\usepackage{multirow}
\usepackage{booktabs}

\usepackage{algorithm}
\usepackage{algpseudocode}

\algrenewcommand\algorithmicrequire{\textbf{Input:}}
\algrenewcommand\algorithmicensure{\textbf{Output:}}

\usepackage[T1]{fontenc}

\usepackage[utf8]{inputenc}

\usepackage{amsmath}
\usepackage{microtype}

\usepackage{inconsolata}

\title{Fixed and Adaptive Simultaneous Machine Translation Strategies Using Adapters}

\author{Abderrahmane Issam \\ {\bf Yusuf Can Semerci} \\ {\bf Jan Scholtes} \\ {\bf Gerasimos Spanakis} \\
        Department of Advanced Computing Sciences \\ 
        Maastricht University \\ 
        \small{\texttt{\{abderrahmane.issam, y.semerci, j.scholtes, jerry.spanakis\}@maastrichtuniversity.nl}}}

\begin{document}
\maketitle
\begin{abstract}
Simultaneous machine translation aims at solving the task of real-time translation by starting to translate before consuming the full input, which poses challenges in terms of balancing quality and latency of the translation. The wait-$k$ policy offers a solution by starting to translate after consuming $k$ words, where the choice of the number $k$ directly affects the latency and quality. In applications where we seek to keep the choice over latency and quality at inference, the wait-$k$ policy obliges us to train more than one model. In this paper, we address the challenge of building one model that can fulfil multiple latency levels and we achieve this by introducing lightweight adapter modules into the decoder. The adapters are trained to be specialized for different wait-$k$ values and compared to other techniques they offer more flexibility to allow for reaping the benefits of parameter sharing and minimizing interference. Additionally, we show that by combining with an adaptive strategy, we can further improve the results. Experiments on two language directions show that our method outperforms or competes with other strong baselines on most latency values. \footnote{Code is available at: \url{https://github.com/issam9/Adapters-SiMT}}

\end{abstract}

\section{Introduction}
Simultaneous machine translation (SiMT) aims at reducing the latency of translation systems. In scenarios with low latency demands, such as conferences or lectures, translating with minimum delay is crucial. In order to reduce the latency, SiMT models start translating before consuming the full input sentence, which improves the latency but affects the quality of the translation, 
because of limited access to enough source context to make a correct prediction. 
SiMT techniques design a strategy to decide when to make a READ (i.e. wait for more source tokens) or WRITE (i.e. output a new token) action. The strategy has to balance the trade-off between quality and latency by making more READ or WRITE actions. Making more READ actions will lead to improved quality but will hinder the latency, while the opposite is true for making more WRITE actions. Fixed policies design a strategy that is detached from whether there is sufficient context to make a WRITE action \cite{ma-etal-2019-stacl, elbayad20_interspeech, zhang-feng-2021-universal}. For instance, the wait-$k$ policy \cite{ma-etal-2019-stacl} trains the model to make $k$ number of READ actions before every WRITE action. The value of $k$ has a direct impact on the quality and latency of the translation and since it is decided during training, wait-$k$ models have to be trained with latency in mind, which means that in order to support multiple latency levels, we need to train multiple models. The multi-path training \cite{elbayad20_interspeech} was introduced to solve this issue by sampling the value of $k$ randomly during training, which results in a model that supports multiple latency levels. This technique was shown to benefit the inference at lower wait-$k$ values by improving the results, but it neglects that parameter sharing between all the wait-$k$ values might introduce interference. \citet{zhang-feng-2021-universal} addressed the interference issue by using Mixture-of-Experts (MoE), where each head of the multi-head attention is treated as an expert and is trained on different wait-$k$ values. This has proven to be a successful technique, but the number of wait-$k$ experts we can introduce depends on the number of heads in the Transformer model, which limits the flexibility in terms of balancing parameter sharing and interference between the wait-$k$ paths. Our method relies on inserting lightweight adapters \cite{rebuffi2017-adapters, pmlr-v97-houlsby19a} for this purpose. The number of the adapters and their capacity can be easily adjusted depending on the wait-$k$ values we intend to support and the complexity of the language direction.

Dynamic strategies have gained increased attention in recent years \cite{gu-etal-2017-learning, zheng-etal-2019-simpler, zheng-etal-2020-simultaneous, Ma2020Monotonic, zhang-feng-2022-information, zhao-etal-2023-adaptive} due to their effectiveness. Dynamic strategies strive to strike a balance between latency and quality by making as much READ actions as necessary and as much WRITE actions as possible. The decision to read or write is made dynamically based on the context (which can be the received input and the previous target tokens) at each decoding step. Although dynamic strategies achieve state-of-the-art results, they often require specialized training techniques \cite{gu-etal-2017-learning, Ma2020Monotonic, zhang-feng-2022-information} that can balance between latency and quality when generating READ/WRITE actions, or even require the training of multiple models \cite{zheng-etal-2020-simultaneous, Ma2020Monotonic} to support multiple latency levels. In order to take advantage of the dynamic wait-$k$ strategies, we adopt a strategy that composes multiple wait-$k$ models during inference (we refer to this as Adaptive Wait-$k$ \cite{zheng-etal-2020-simultaneous}) to work with wait-$k$ adapters instead. This brings efficiency and cost benefits as only one model is required to satisfy multiple latency levels and also improves performance compared to other strong baselines including Adaptive Wait-$k$.

In summary, our main contributions are the following:
\begin{itemize}
    \item We introduce lightweight adapters as a flexible solution to balance parameter sharing and interference in multi-path training.
    \item We show that by combining adapters with a simple adaptive strategy (i.e. Adaptive Wait-$k$) we can further improve the results.
    \item We show that our technique outperforms or competes with other strong baselines on most latency levels.
\end{itemize}

\section{Related Works}

\subsection{Adapters for Machine Translation}

Adapters \cite{rebuffi2017-adapters, pmlr-v97-houlsby19a} are typically small modules that are used in order to efficiently adapt a pre-trained model to a downstream task, where the pre-trained model can be either frozen \cite{pmlr-v97-houlsby19a}, or trained jointly with the adapters \cite{pmlr-v97-stickland19a}. 

Adapters have been used for efficient multi-task fine-tuning \cite{pmlr-v97-stickland19a}, where each set of adapters is trained on a specific task. \citet{pfeiffer-etal-2021-adapterfusion} added AdapterFusion on top of the adapters as a way to compose the representations of different tasks. \citet{pfeiffer-etal-2022-lifting} used adapters as language-specific parameters in order to address the curse of multilinguality in multilingual pre-training, where the adapter modules are introduced during pre-training instead of post-hoc. 

For Neural Machine Translation (NMT), \citet{bapna-firat-2019-simple} introduced a simple formulation of adapters to learn language-pair specific parameters, where they showed that it improves performance on high resource languages in Multilingual Translation. \citet{chronopoulou-etal-2023-language} trained language-family adapters to address negative interference while allowing for parameter sharing between similar languages, which improved performance on low resource languages. \citet{zhao-calapodescu-2022-multimodal} fine-tuned adapters on multimodal noise, then added a fusion layer in order to improve generalization to other types of noise. Adapters were also explored for other motivations like Zero-shot NMT and unsupervised domain adaptation \cite{philip-etal-2020-monolingual, malik-etal-2023-udapter}.

\subsection{Simultaneous Machine Translation}
SiMT systems can be divided into fixed and adaptive policies. Fixed policies rely on predefined rules for READ/WRITE decisions. \citet{ma-etal-2019-stacl} proposed the wait-$k$ policy, where the model starts by reading $k$ tokens then alternates between reading and writing one token. \citet{elbayad20_interspeech} introduced multi-path training, where one model is trained to support multiple wait-$k$ values by sampling $k$ randomly during training. \citet{zhang-feng-2021-universal} addressed interference in multi-path training by using Mixture-of-Experts. \citet{Zhang_Feng_Li_2021} used Knowledge Distillation from a Full-Sentence Transformer to embed future information into the SiMT model. For adaptive policies, \citet{gu-etal-2017-learning} trained a Reinforcement Learning agent to decide READ/WRITE actions, where the reward function is designed to consider both quality and latency. \citet{zheng-etal-2019-simpler} generated supervised READ/WRITE actions then trained a classification model to predict the action based on encoder and decoder representations. \citet{zheng-etal-2020-simultaneous} introduced a heuristic strategy to compose wait-$k$ models into an adaptive policy based on their uncertainty. \citet{zhang-zhang-2020-dynamic} trained a sentence segmentation model to predict complete sentences and feed them through a full-sentence translation model. \citet{arivazhagan-etal-2019-monotonic} introduced MILK, where they modified the attention mechanism to learn a Bernoulli variable to decide READ/WRITE actions. \citet{Ma2020Monotonic} adapted MILK to the transformer architecture. \citet{zhang-feng-2022-information} proposed ITST, which quantifies the transported information from source to target then generates a token when the quantity is deemed sufficient. \citet{zhao-etal-2023-adaptive} trained a supervised policy network based on automatically generated divergence between the predicted distribution of partial and full sentence input.

The majority of the techniques outlined require training multiple models to accommodate different latency levels. Our approach focuses on the efficient training of a single model that can support various latency levels at inference time.

\section{Background}

\subsection{Adapters}
Adapters are lightweight modules that can be inserted into a model for the purpose of task or domain adaptation \cite{pmlr-v97-houlsby19a, bapna-firat-2019-simple}. They offer an efficient solution for fine-tuning the model and limiting catastrophic forgetting \cite{pmlr-v97-houlsby19a}. 

Formally, for a set of $N$ tasks and a model $M$, the adapter parameters $A$ are introduced. We assume that for each task we have a dataset $D_n$. The model parameters can be frozen or jointly trained with the adapters. For a frozen model, the model $M$ is pre-trained and the objective function for task $n \in \{1,...,N\}$ can be defined as:
\begin{equation}
A_n \leftarrow \underset{A_n}{argmin} \, L_n(D_n;M, A_n)   
\end{equation}
The parameters $A_n$ are randomly initialized for each task, then they are trained on the dataset $D_n$ in order to minimize the loss function $L_n$. This results in $N$ adapters that can specialize the model representations to each task $n$.

In the case of jointly training the model and the adapters, the model parameters $M$ can be randomly initialized or frozen. The objective function can be defined as:
\begin{equation}
M' \leftarrow \underset{M, A}{argmin} \, \left ( \sum_{n=1}^{N} L_n(D_n;M, A_n) \right )
\end{equation}
where $M'$ is both the parameters of the model $M$ and the adapters $A_n$ for $n \in \{1, ..., N\}$. The parameters $A_n$ are activated during training depending on the task $n$.

\subsection{Wait-$k$ Policy}
The wait-$k$ policy \cite{ma-etal-2019-stacl} trains a model to start translating after receiving $k$ source tokens. The model then alternates between writing and reading a new token. It is a fixed policy, where the $k$ value has to be chosen during training and inference. The model reads $g_k(t)$ number of source tokens from the source sentence $x=(x_1, ..., x_m)$ when generating the target token $y_t$, where $g_k(t)$ is defined as:
\begin{equation}
    \label{eq_gt}
    g_k(t) = min\{|x|, t+k-1\}
\end{equation}

Instead of training the model for a specific wait-$k$ value, \citet{elbayad20_interspeech} introduced the multi-path training, which samples $k$ uniformly from $[1,...,|x|]$ for each batch during training. This enables the model to support multiple wait-$k$ values and allows for information sharing between different wait-$k$ paths. While it was shown that the multi-path training improves the results over the wait-k policy, it does not offer a solution to balance between parameter sharing and interference that we aim at solving by introducing adapters.

\section{Method}
Our method is composed of two steps: first we train a single model that can support multiple fixed wait-$k$ values by using wait-$k$ adapters, then we rely on the probability that the model assigns to the most likely token in order to build an adaptive strategy, where we decide a READ or WRITE action based on a predefined probability threshold.

\subsection{Multi-path Training with Adapters}
\label{sec4.1}
\begin{figure*}[ht]
  \centering
\includegraphics[scale=0.23]{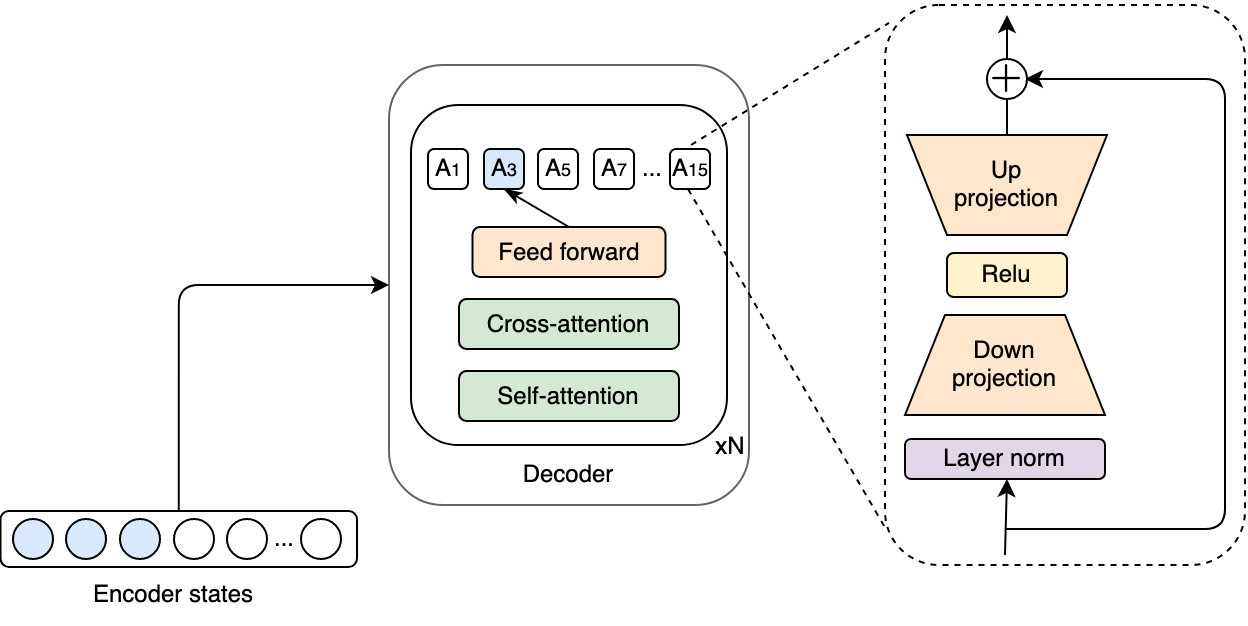}
  \caption{Transformer Decoder with Adapters Wait-$k$, we illustrate an example where 8 adapters are inserted with $K_A=\{1,3,5,7,9,11,13,15\}$, the generation step is $t=0$, and $A_3$ is activated because $k=3$.}
  \label{fig_adapters_waitk}
\end{figure*}

Multi-path training is highly advantageous as an efficient alternative to the wait-$k$ policy, where we need to train multiple models to support more than one latency at inference, but might introduce interference between wait-$k$ paths due to parameter sharing. In order to provide the ability to balance between parameter sharing and interference, we introduce adapters into each decoder layer and we activate adapters according to the wait-$k$ paths they are meant to support. Figure \ref{fig_adapters_waitk} shows an illustration of this. During training, the wait-$k$ value for each batch is sampled uniformly from $[1,...,|x|]$ following the multi-path training \cite{elbayad20_interspeech} and based on that, the model decides which adapter will be activated. We set the adapter lagging $K_A$ as a list of equally spaced positive integers in increasing order, where each integer specifies the minimum wait-$k$ value supported by each adapter. We insert one adapter for each value in $K_A$. Since the train wait-$k$ is randomly sampled from $[1, …, |x|]$, we train each adapter on values starting from its minimum wait-$k$ up until the minimum wait-$k$ of the next adapter. For example, we can set $K_A = \{1, 5, 9, 13\}$ and this will indicate adding 4 adapters, where each adapter will handle 4 wait-$k$ values (starting from each integer in $K_A$ until the next), except the fourth adapter ($k_A$ = 13), which will handle values starting from 13 up until the length of the input sequence $|x|$. We follow \citet{bapna-firat-2019-simple} implementation and insert the residual adapter modules after the feed-forward layer. Algorithm \ref{alg1} shows the pseudo-code for computing the decoder hidden states at decoding step $t$ using Adapters Wait-k, where $H^0$ is considered to be the input embeddings of the decoder, and $g_{k}(t)$ is computed based on equation \ref{eq_gt}.  

\begin{algorithm}
\caption{Adapters Wait-$k$ Policy}
\label{alg1}
\label{alg:adapter-waitk}
\begin{algorithmic}
\Require Encoder output $Z$, Decoder hidden states $H_t$, Adapter lagging $K_{A}$, Test lagging $k_{\text{test}}$
\Ensure Hidden states $H_t^L$
\If{is\_training}
    \State $k \gets$ Sample from $[1, \ldots, |Z|]$
\Else
    \State $k \gets k_{\text{test}}$
\EndIf
\For{$k_{A}$ in $K_{A}$}
    \If{$k \geq k_{A}$}
        \State $A^l = A_{k_{A}}^l$   \hspace{1cm} for $l \in [1, \dots, L]$
    \EndIf
\EndFor
\For{$l \gets 1$ to $L$}
    \State $H_t^l = Decoder^l(H_t^{l-1}, Z_{\leq g_k(t)})$
    \State $H_t^l = A^l( H_t^l ) + H_t^l $
\EndFor
\State \textbf{Return} $H_t^L$
\end{algorithmic}
\end{algorithm}

\subsection{Adaptive Adapters}
\label{sec4.2}
We follow \citet{zheng-etal-2020-simultaneous} to build an adaptive strategy by using adapters instead of different models for each wait-$k$ value, which can be computationally expensive and less efficient. At each decoding step, we activate one adapter based on the lagging behind the current generation step, which is calculated as $k = |x| - |y|$, where $|x|$ is the number of input tokens and $|y|$ is the number of generated tokens. At the beginning of generation, $|x|=1$ and $|y|=0$, which means $k$ starts from $1$. Then, we rely on the probability of the most likely token to decide whether to write or read a new token. If the probability is less than a threshold $\rho_k$, we read a new token, otherwise, we write. The possible values of $k$ are between $k_{min}$ and $k_{max}$ that we determine during inference. If $k$ is lower than $k_{min}$, we force the model to read, if it is higher or equal to $k_{max}$, we force the model to write, which means that the choice of $k_{min}$ and $k_{max}$ also impacts the trade-off between latency and quality (as we analyze in Section \ref{sec6.1}). When the whole input sequence is consumed (i.e. $x_{|x|}=\text{</s>}$), we set $k$ to $k_{max}$ and generate the rest of the target sequence. Algorithm \ref{alg2} shows the pseudo-code of this method using adapters. 

\begin{algorithm}
\caption{Uncertainty based Adaptive Policy}
\label{alg2}
\begin{algorithmic}
\Require Two integers $k_{\text{min}}$ and $k_{\text{max}}$ and a sequence of thresholds $\rho_k$ for $k_{\text{min}} \leq k \leq k_{\text{max}}$.
\Ensure Predicted sequence $y$
    \While{$x_{|x|} \neq \text{</s>}$ and $y_{|y|} \neq \text{</s>}$}
        \State $k \gets |x| - |y|$
        \If {$k < k_{\text{min}}$}
            \State $x \gets x \circ \text{READ()}$ \Comment{READ action}
        \Else
            \State $y_{\text{top}}, p_{\text{top}} \gets P_k(M, A_k, x, y)$
            \If{$k \textless k_{\text{max}}$ and $p_{\text{top}} \textless \rho_k$}
                \State $x \gets x \circ \text{READ()}$ \Comment{READ action}
            \Else
                \State $y \gets y \circ y_{\text{top}}$ \Comment{WRITE action}
            \EndIf
        \EndIf
    \EndWhile
\While{$y_{|y|} \neq \text{</s>}$}
    \State $y_{\text{top}}, p_{\text{top}} \gets P_{k_{\text{max}}}(M, A_{k_{\text{max}}}, x, y)$
    \State $y \gets y \circ y_{\text{top}}$ \Comment{WRITE action}
\EndWhile
\State \Return $y$
\end{algorithmic}
\end{algorithm}

\section{Experiments}
In this section, we describe the datasets we used to evaluate the models and the baselines that we compare against along with the evaluation setup. We also provide the main results of our experiments.

\subsection{Datasets}
We evaluate our method on two public datasets: the En-Vi dataset for Transformer-Small and De-En for both Transformer-Base and Transformer-Big.

\textbf{IWSLT15\footnote{\url{nlp.stanford.edu/projects/nmt/}}} \textbf{English} $\rightarrow$ \textbf{Vietnamese} (133K pairs) \cite{cettolo-etal-2015-iwslt}. We follow the settings of \citet{pmlr-v70-raffel17a} and \citet{Ma2020Monotonic}. We use TED tst2012 (1553 pairs) as the validation set and TED tst2013 (1268 pairs) as the test set. We replace tokens with frequency less than 5 with $<unk>$. The final vocabulary sizes are 17K and 7.7K for English and Vietnamese respectively.

\textbf{WMT15\footnote{\url{www.statmt.org/wmt15/}}} \textbf{German} $\rightarrow$ \textbf{English} (4.5M pairs) We follow the settings of \citet{ma-etal-2019-stacl}. We use newstest2013 (3000 pairs) as the validation set and newstest2015 (2169 pairs) as the test set. We apply BPE \cite{sennrich-etal-2016-neural} with 32K merge operations jointly on the source and target to construct a shared vocabulary.

\subsection{System Settings}
\label{sec5.2}
We conduct experiments on the following systems:

\textbf{Full Sentence:} \cite{vaswani-transformer}
Standard Transformer model that takes the full sentence as input before starting to translate.

\textbf{Wait-$k$:} \cite{ma-etal-2019-stacl}
A simple policy that waits for $k$ source tokens before starting to alternate between writing a target token and reading a source token.

\textbf{Multi-path Wait-$k$:} \cite{elbayad20_interspeech}
Trains a model to support multiple wait-$k$ policies by randomly sampling $k$ during training, then the $k$ value is fixed during inference.

\textbf{Adaptive Wait-$k$:} \cite{zheng-etal-2020-simultaneous}
It is a method for composing multiple wait-$k$ models during inference in order to build an adaptive strategy. The model is selected based on the lagging behind the generation step, and the decision to write or read is based on the output probabilities.

\textbf{MoE Wait-$k$:} \cite{zhang-feng-2021-universal} Mixture-of-Experts Wait-$k$ is similar to Multipath Wait-$k$ but applies experts to learn different wait-$k$ policies to avoid interference.

\textbf{MMA:} \cite{Ma2020Monotonic} Monotonic multi-head attention (MMA) jointly learns a Bernoulli variable that is used to decide READ/WRITE action.

\textbf{Adapters Wait-$k$:} Our method as described in Section \ref{sec4.1}.

\textbf{Adaptive Adapters:} Our method as described in Section \ref{sec4.2}.

All implementations are based on the original Transformer architecture \cite{vaswani-transformer} and are using the Fairseq library \cite{ott-etal-2019-fairseq}. We apply Transformer-Small (4 heads) for En-Vi and both Transformer-Base (8 heads) and Transformer-Big (16 heads) for De-En. The encoder is made unidirectional to avoid encoding the source input each time a new token is added.

The evaluation is performed using BLEU \cite{papineni-bleu} for translation quality and Average Lagging (AL)\footnote{\url{github.com/SimulTrans-demo/STACL}} \cite{ma-etal-2019-stacl} for latency. AL measures by how many tokens the system is lagging behind an ideal policy (a wait-$k$ policy with $k=0$). Given $g(t)$, AL is computed as:

\begin{equation}
A\!L_g(x, y) = \frac{1}{\tau_g(|x|)} \sum_{t=1}^{\tau_g(|x|)} g(t) - \frac{(t - 1)}{|y|/|x|}
\end{equation}
where $x$ and $y$ are the source and target sentences respectively, while $\tau_g(|x|) = \min \{ t \,|\, g(t) = |x| \}$ is the decoding step where the source sentence finishes.

We set the adapter lagging to $K_A = \{1, 3, 5, 7, 9, 11, 13, 15\}$ for our experiments, which means that 8 adapters are inserted into the model and we specify the adapter bottleneck size as 64. In Table \ref{num_params_table}, we report the number of parameters of each method and the number of models required to achieve the latency levels reported in the results section. Adapters Wait-$k$ policy introduces 79.94M parameters into Transformer-Big, but still has the advantage of using one model to support multiple latency levels. In Section \ref{sec6.2}, we experiment with other settings of $K_A$ in order to shed light on how much sharing is best between wait-$k$ values during the multi-path training.

\begin{table}[ht]
\centering
\begin{tabular}{|l|c|c|c|}
\hline
\textbf{Model} & \textbf{\#Parameters} & \textbf{\#Models} \\
\hline
Full Sentence & 209.91M & 1 \\
Wait-$k$ & 209.91M & 5 \\
Adaptive Wait-$k$ & 209.91M & 13 \\
Multipath & 209.91M & 1 \\
MMA & 222.51M & 7 \\
MoE Wait-$k$ & 209.91M & 1 \\
Adapters Wait-$k$ & 289.85M & 1 \\
Adaptive Adapters & 289.85M & 1 \\
\hline
\end{tabular}
\caption{The number of parameters of the models for Transformer-Big on De-En along with the number of models required to achieve different latency levels.}
\label{num_params_table}
\end{table}

The adaptive strategy requires three parameters to be specified at inference, namely, $k_{min}, k_{max}$, and the probability threshold $\rho_k$. For En-Vi experiments, $k_{min}$ and $k_{max}$ are set to $1$ and $9$ respectively, while for De-En, we lower $k_{max}$ to $5$, which we have found to improve the results in low latency. We analyze this effect in Section \ref{sec6.1}. $\rho_k$ decreases as a function of the lagging $k$, since we want the model to be more aggressive when $k$ is low and more conservative when $k$ is high. We set $\rho_{k_{min}}$ and $\rho_{k_{max}}$ and compute the threshold as: $\rho_k = \rho_{k_{min}} - d.(k-1)$, where $k_{min} \leq k \leq k_{max}$ and $d=(\rho_{k_{min}} - \rho_{k_{max}})/(k_{max}-k_{min})$. In order to vary the latency, we test the following values of $\rho_{k_{min}}$ and $\rho_{k_{max}}$: $\rho_{k_{min}} \in \{0.2, 0.4, 0.6, 0.8, 1.\}$, $\rho_{k_{max}}$ = $0.$, and $\rho_{k_{min}}$ = $1.$, $\rho_{k_{max}} \in \{0.2, 0.4,0.6, 0.8\}$.

\subsection{Main Results}

\begin{figure*}[ht]
  \centering
  \subfigure[En-Vi, Transformer-Small]{\includegraphics[scale=0.4]{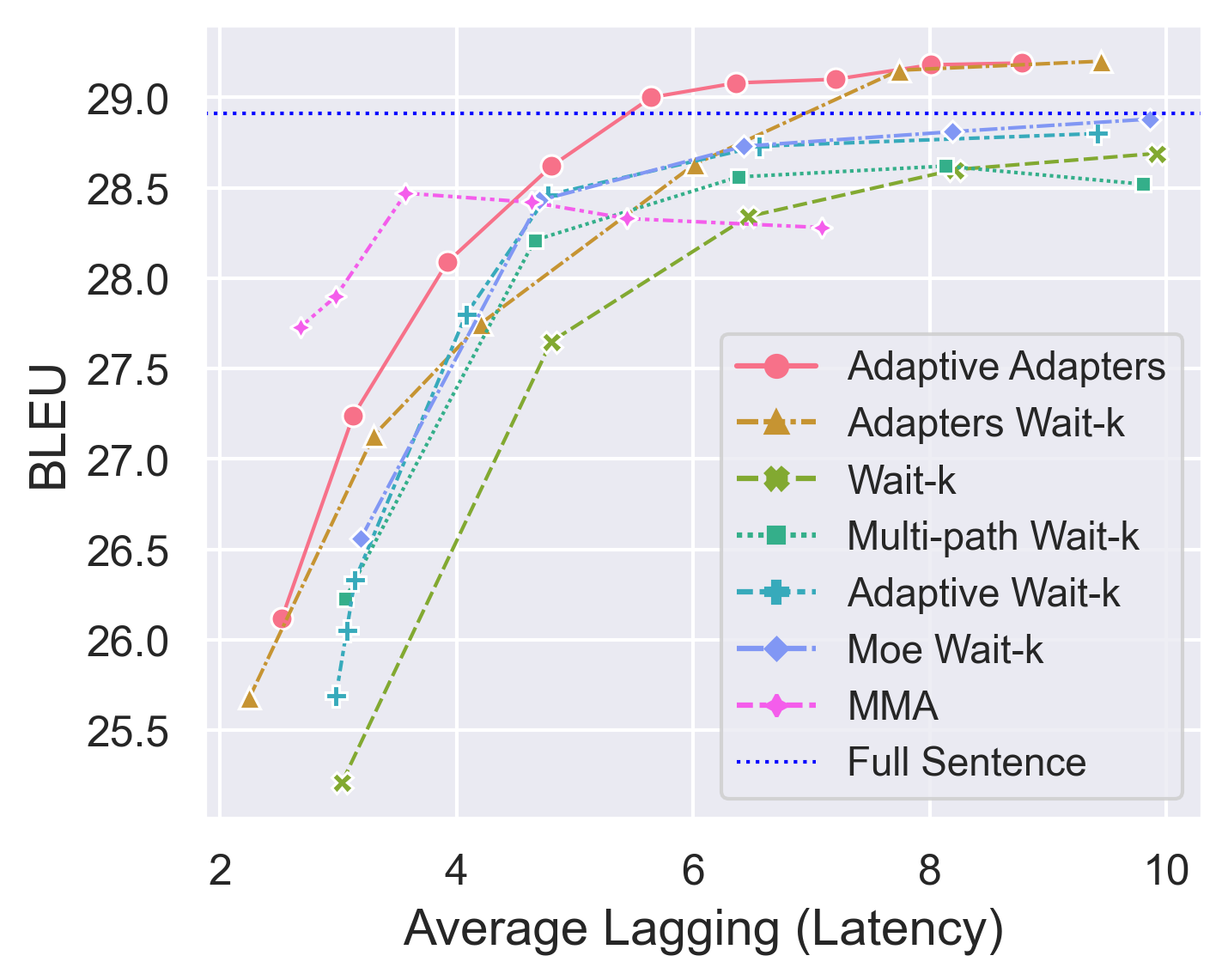}
  \label{en-vi-results}}
  \subfigure[De-En, Transformer-Base]{\includegraphics[scale=0.4]{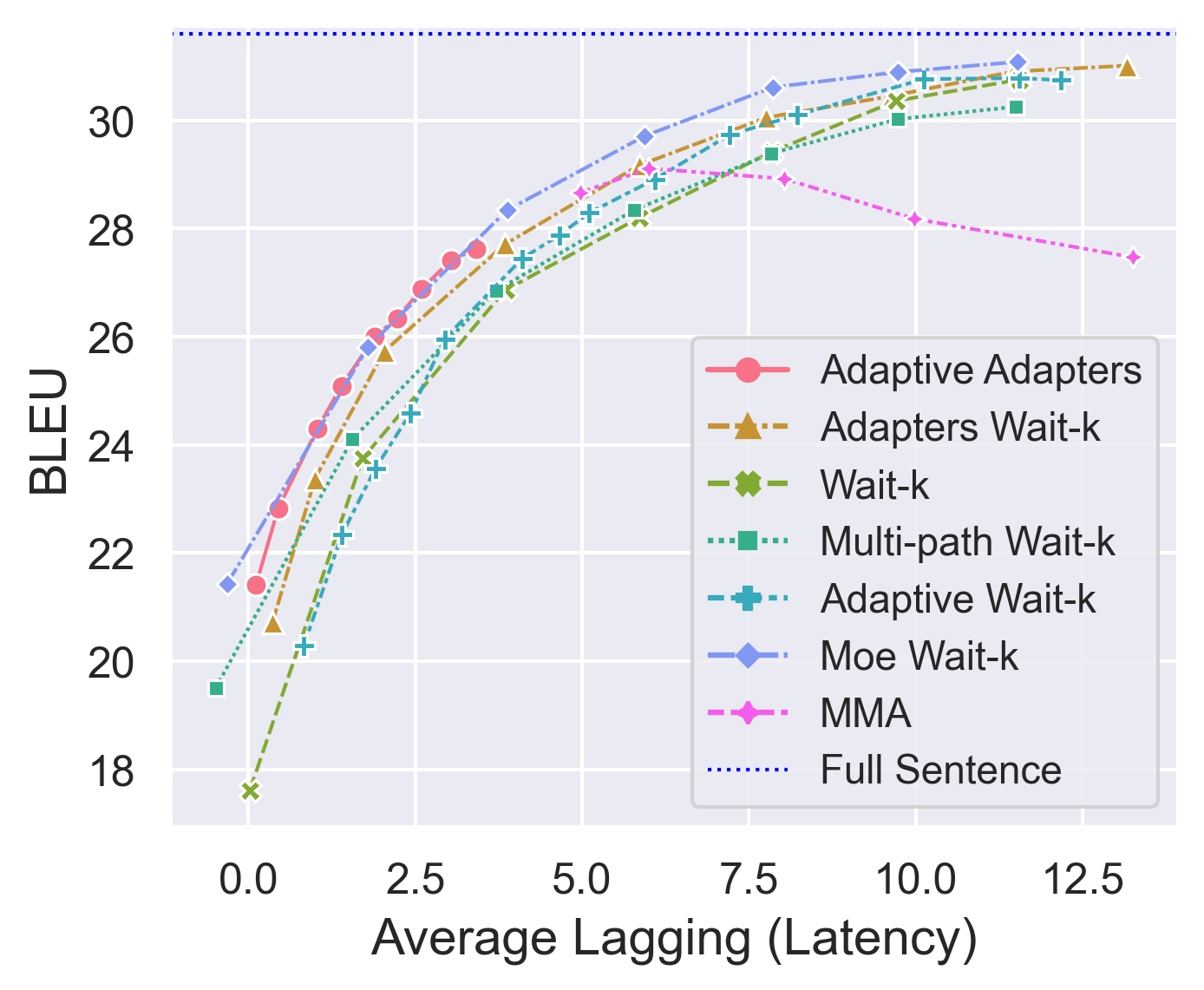}
  \label{de-en-results}}
  \subfigure[De-En, Transformer-Big]{\includegraphics[scale=0.4]{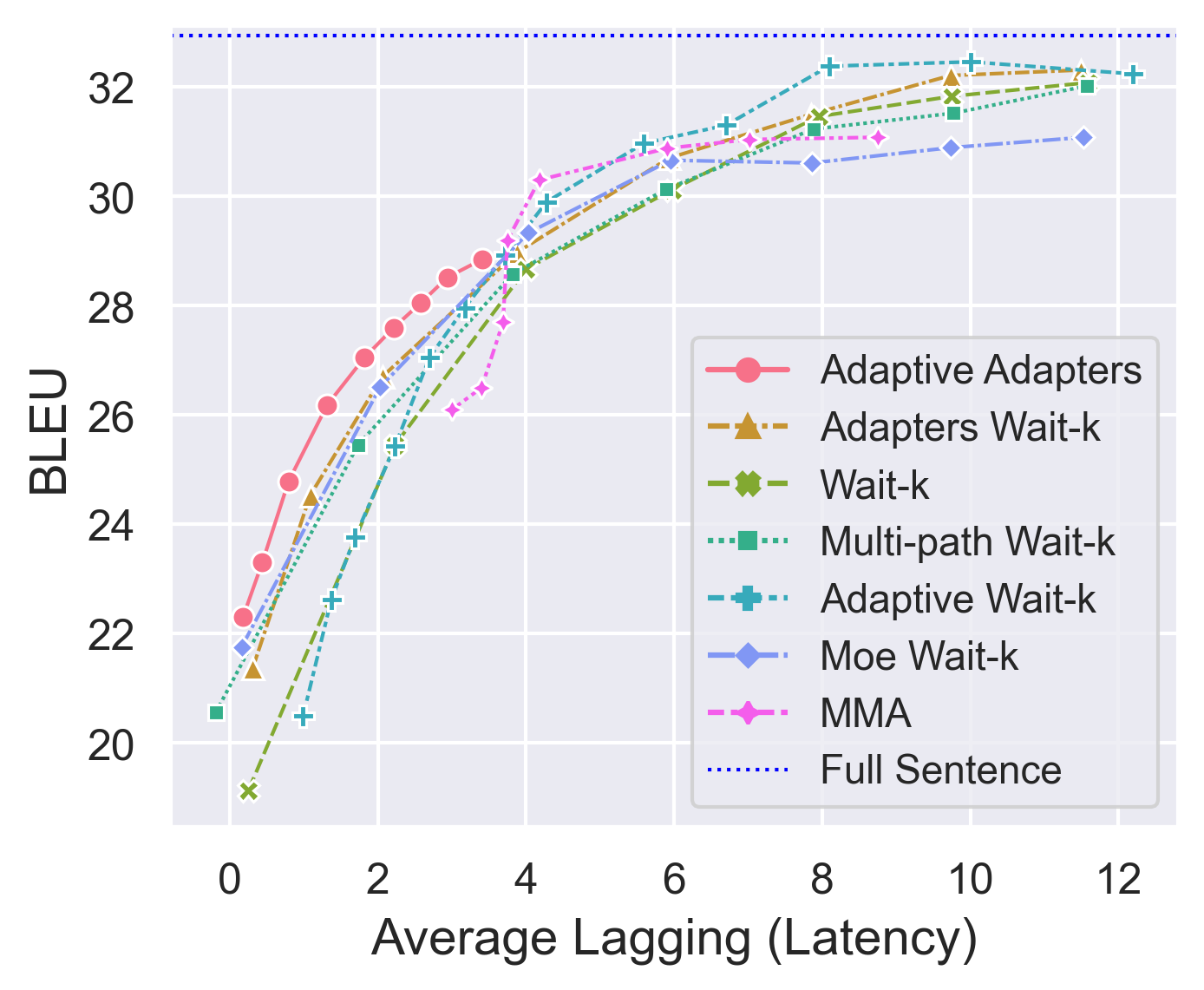}
  \label{de-en-big-results}}

  \caption{Translation quality (BLEU) against latency (AL) of our methods (Adaptive Adapters, Adapters Wait-$k$) and previous adaptive (MMA, Adaptive Wait-$k$) and fixed (Wait-$k$, MoE Wait-$k$, Multi-path Wait-k) strategies on En-Vi and De-En.}
  \label{fig_main_results}
\end{figure*}

In Figure \ref{fig_main_results}, we compare our methods to previous adaptive and fixed strategies on two language directions. We find that our method improves or competes with other strategies while using a single model. MMA, Wait-$k$, and Adaptive Wait-$k$ require the training of multiple models in order to support different latency levels (as seen in Table \ref{num_params_table}), while our method is more efficient in this regard. Adapters Wait-$k$ is competitive with other strong fixed strategies like MoE Wait-$k$ and Multi-path Wait-k and it brings further improvements to combine it with the adaptive strategy. 

Our method does not support higher latency on De-En because we are using a $k_{max}$ value of $5$ (as seen in Figures \ref{de-en-results} and \ref{de-en-big-results}), which we have found to improve results for low latency. However, we show the results for higher $k_{max}$ and compare them with Adaptive Wait-k on De-En in Section \ref{sec6.1}.
 
Using adapters alone is competitive with other methods, especially on En-Vi (as seen as in Figure \ref{en-vi-results}). Compared to Multi-path Wait-k, our method achieves better results on most latency levels, which shows the importance of minimizing interference between different lagging values. Combining our method with an adaptive strategy further improves the results, especially in low latency. In comparison to Adaptive Wait-$k$, where wait-$k$ policy models are trained and composed during inference, we find that our method is better in all latency levels while being more efficient.

Compared to MoE Wait-$k$, which also aims at minimizing interference introduced by multi-path training \cite{zhang-feng-2021-universal}, we find that our method is better in all latency levels on En-Vi and De-En with Transformer-Big (as seen in Figures \ref{en-vi-results} and \ref{de-en-big-results}), while achieving competitive results when using Transformer-Base (as seen in Figure \ref{de-en-results}). Our method is more flexible in terms of balancing the trade-off between parameter sharing and interference, as we can choose the number of wait-$k$ values supported by each adapter and we can also manipulate the capacity of the adapters by adjusting the bottleneck size. This can bring further improvements but requires more experimentation to find the appropriate hyperparameters.

\section{Analysis}
In this section, we look into how the performance changes in response to varying the value of $k_{max}$, then we provide a wall-clock time comparison between Adapters Wait-$k$ and Multi-path Wait-$k$. Moreover, we experiment with how balancing between parameter sharing and interference by adjusting the adapter lagging impacts the performance, and also experiment with varying the bottleneck size in order to discern the impact of the complexity of the adapters. At last, we analyze the L2-norm of the adapter representations to discover which adapter layers are involved in the prediction.

\subsection{Ablation}
\label{sec6.1}
\begin{figure}[H]
  \centering
\includegraphics[scale=0.53]{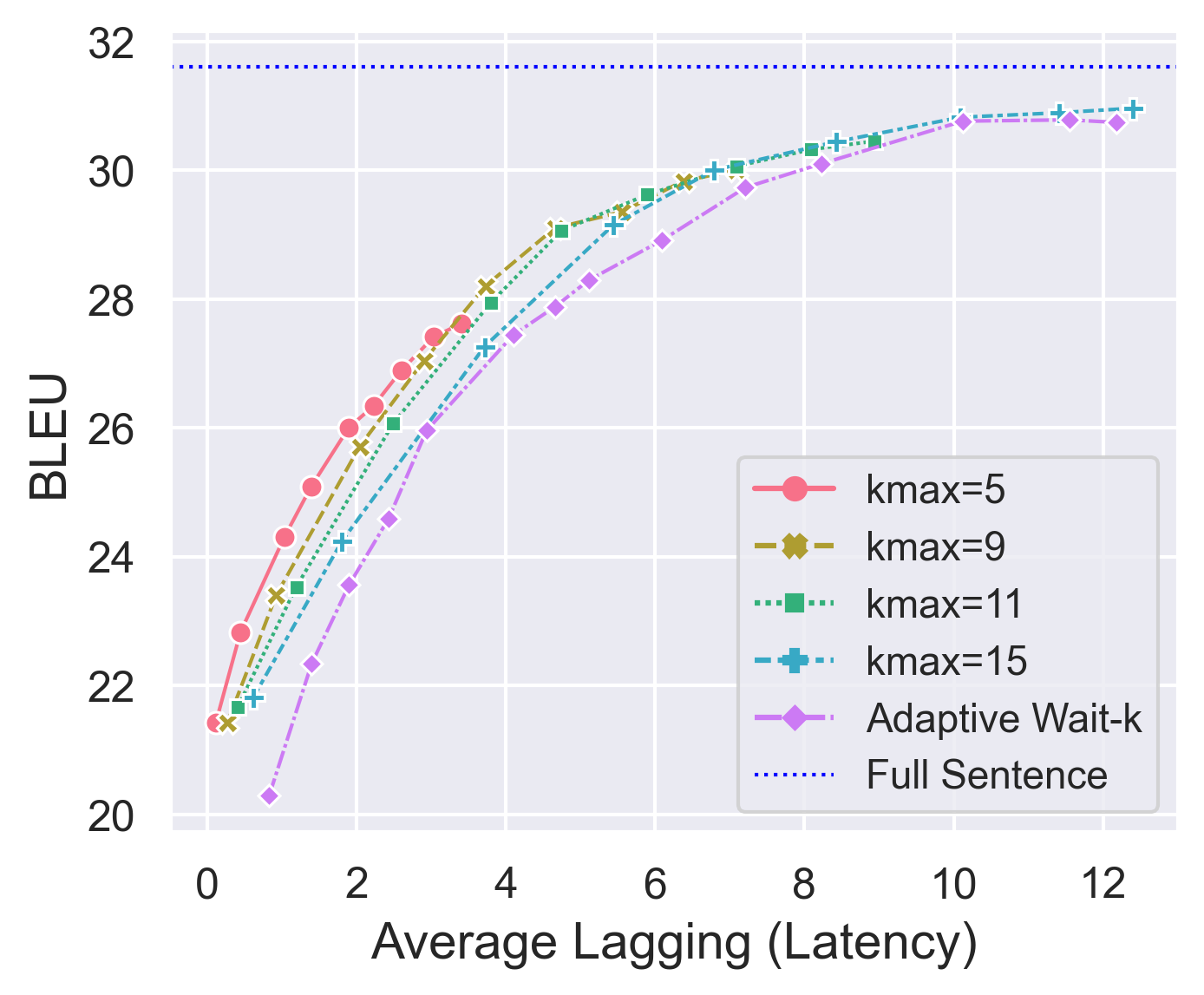}
  \caption{Results of increasing the value of $k_{max}$ on De-En. Lower $k_{max}$ values achieve better BLEU score in low latency, but it is necessary to increase the value of $k_{max}$ in order to support high latency.}
\label{fig_kmax}
\end{figure}

We found that lowering the value of $k_{max}$ for the adaptive strategy improves the results in low latency, which we believe is the priority in SiMT, but a lower $k_{max}$ value also limits the ability of supporting high latency. In Figure \ref{fig_kmax}, we show that by increasing the value of $k_{max}$ we can support high latency and get better quality translations. We compare to Adaptive Wait-$k$ and show that we still achieve better results for all the values of $k_{max}$. A lower $k_{max}$ forces the model to be more aggressive, which in some cases can improve the results in lower latency. The fact that forcing the model to be more aggressive improves the performance signifies that the adaptive strategy decides to wait in cases where the model is able to make a correct prediction, which suggests that the adaptive strategy based on the probability threshold can still be improved by a better strategy.

\subsection{Inference Time}
\begin{figure}[H]
  \centering
\includegraphics[scale=0.5]{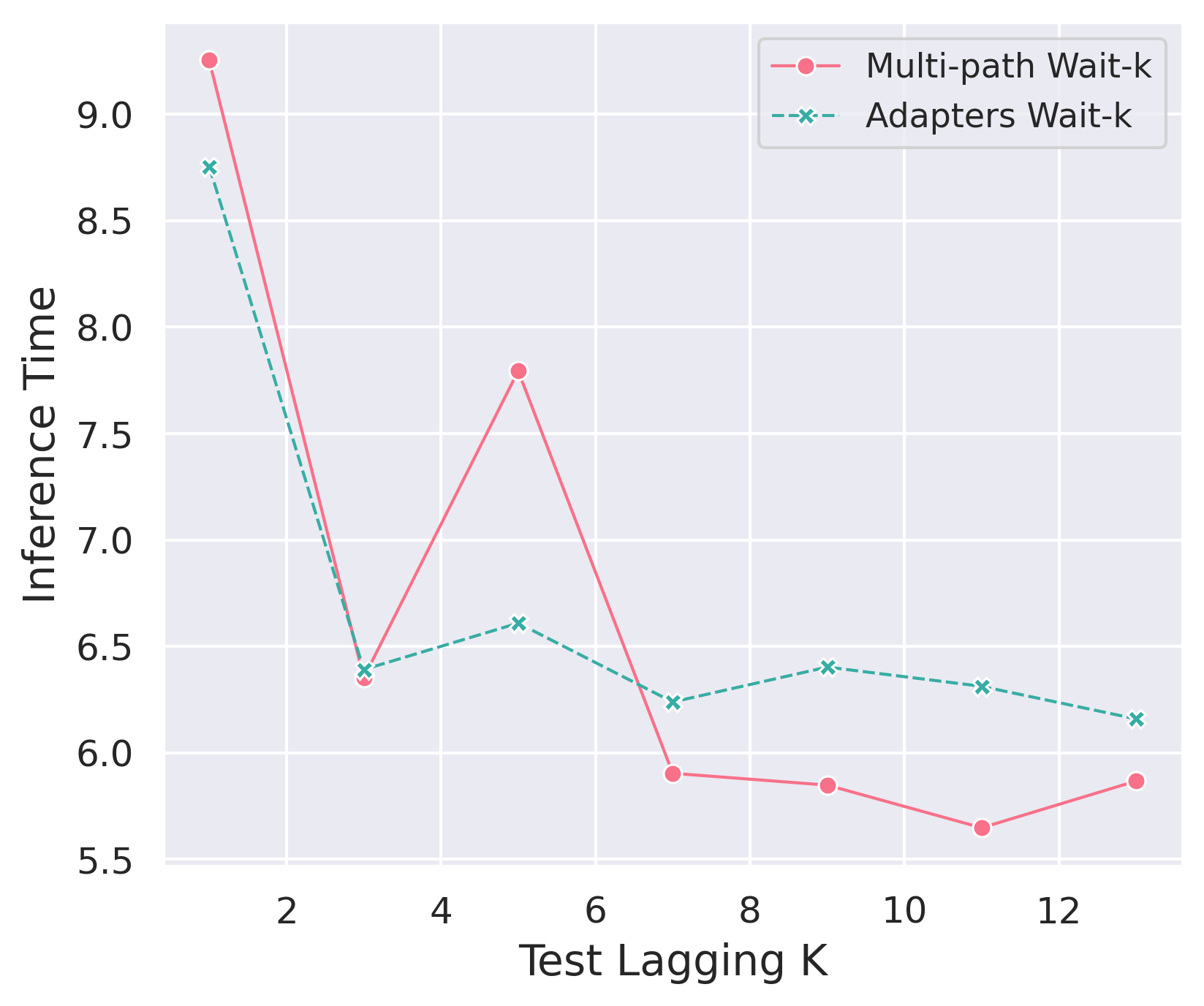}
  \caption{Wall-clock time comparison between Adapters Wait-$k$ and Multi-path Wait-$k$ averaged over 5 runs on En-De.}
\label{inference_time}
\end{figure}
Although our method has more parameters than the baseline Multi-path Wait-$k$ due to the additional adapters, the effect on the inference time is not proportional to the number of adapters because only one adapter is activated at a time. To illustrate this, we compare the wall-clock inference time (averaged over 5 runs) of Adapters Wait-$k$ and Multi-path Wait-$k$ in Figure \ref{inference_time}. It seems that adapters are faster in low $k$ values which could be due to over generation by the Multi-path model (where the model generates longer sequences than it should), while starting from a $k$ value of $7$, Multi-path Wait-$k$ is better and the difference fluctuates between 0.29s and 0.66s.

\subsection{Adapter Lagging}
\label{sec6.2}
\begin{figure}
  \centering
\includegraphics[scale=0.5]{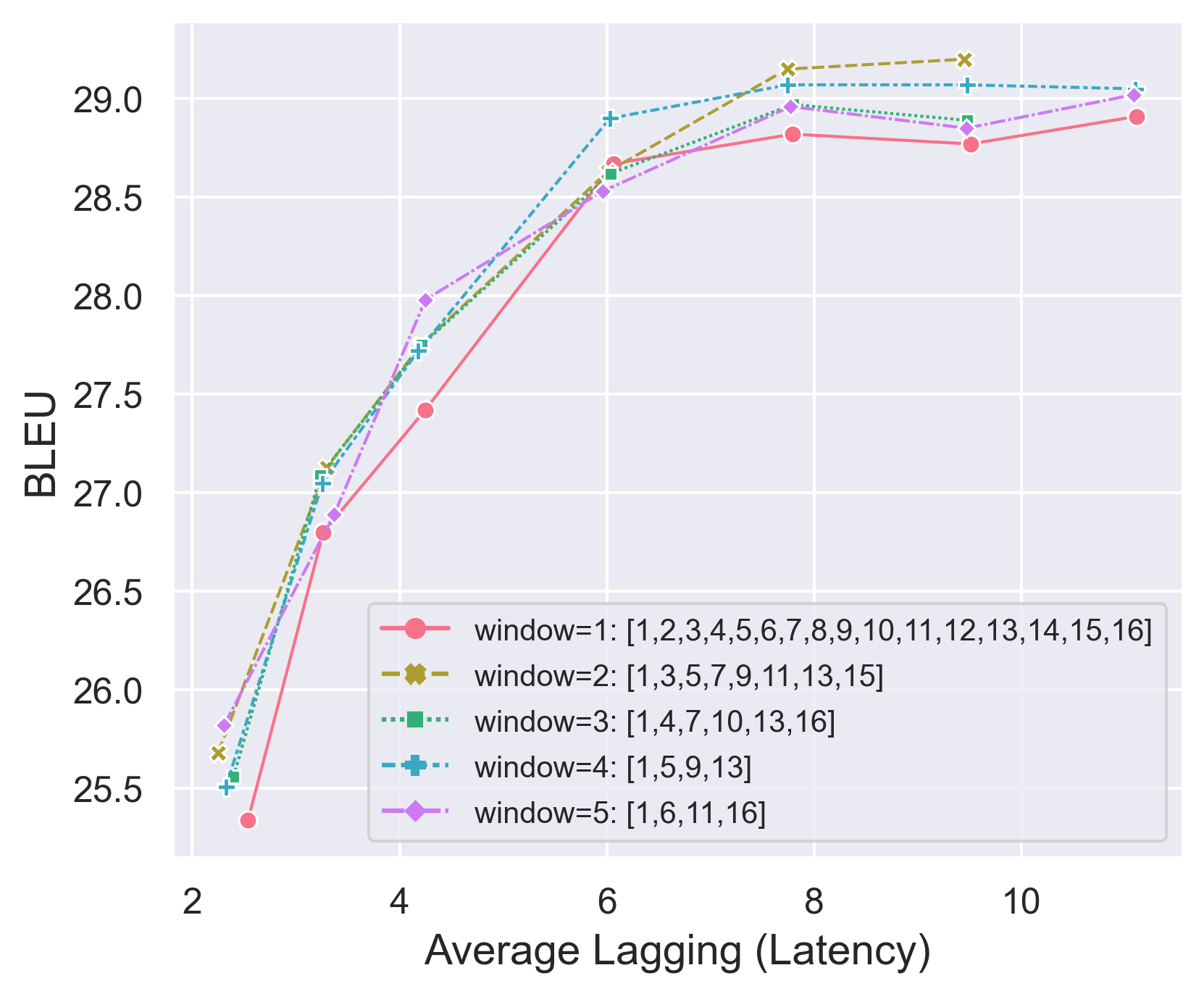}
  \caption{Results of varying the window sizes of the adapter lagging between 1 and 5 on En-Vi.}
\label{fig_adapter_lagging}
\end{figure}

The adapter lagging $K_A$ specifies the number of wait-$k$ values that one single adapter will support and also the number of adapters that we will use. We vary the adapter lagging window between $1$ and $5$, while maintaining the range between $1$ and $16$. The results are shown in Figure \ref{fig_adapter_lagging}. The wait-$k$ values supported by an adapter controls the amount of sharing and interference between the values. For example, for $K_A=\{1,5,9,13\}$, adapter $A_1$ will be trained on $k \in \{1,2,3,4\}$. We note that although it has more parameters, a window of $1$ achieves the worst results, which signifies that parameter sharing between wait-$k$ values is crucial. Adapter lagging with window $4$ and $5$ are competitive especially in low latency, which indicates that lower wait-$k$ values benefit more from sharing. This is consistent with the fact that wait-$k$ models achieve better results when tested on lower wait-$k$ values \cite{zhang-feng-2021-universal}. 

\subsection{Adapter Bottleneck}
\begin{figure}[H]
  \centering
\includegraphics[scale=0.5]{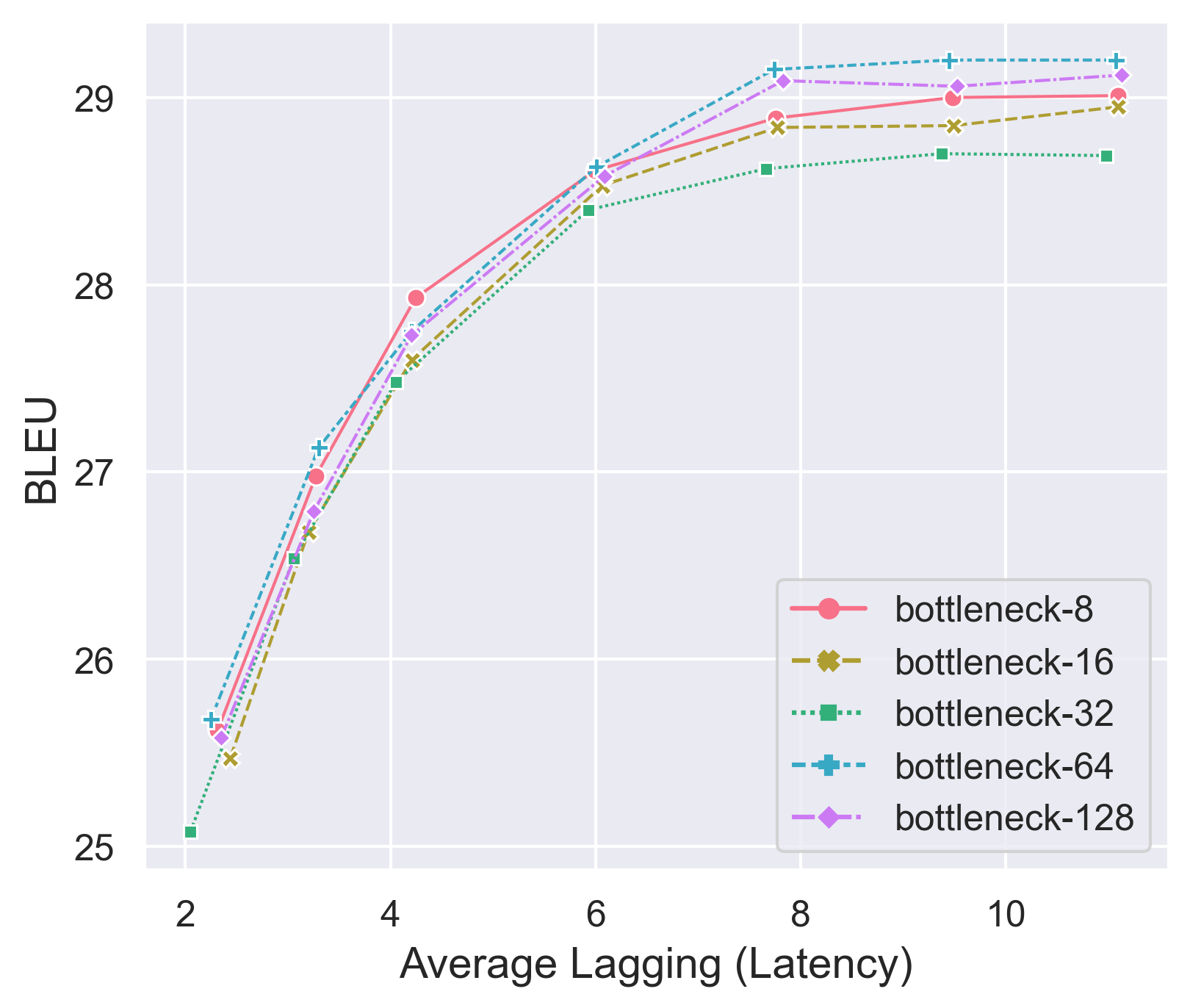}
  \caption{Results of doubling the bottleneck size of the adapters on En-Vi.}
  \label{fig_adapter_bottleneck}
\end{figure}

The adapter's bottleneck size can be used to tune the representation capacity of the adapters and can be interesting to tune depending on the language pair and the adapter lagging. In Figure \ref{fig_adapter_bottleneck}, we experiment with doubling the adapter's bottleneck size from $8$ to $128$, which can be regarded as increasing the representation capacity of the adapter network. We found that the bottleneck size impacts the performance but not in a consistent way - as in larger size results in better performance - but it seems to interact with other hyperparameters (e.g. adapter lagging) to improve or hinder the performance, especially in high latency, where the gap in performance is larger.

\subsection{Adapter Representation Norm}
\begin{figure}[H]
  \centering
\includegraphics[scale=0.5]{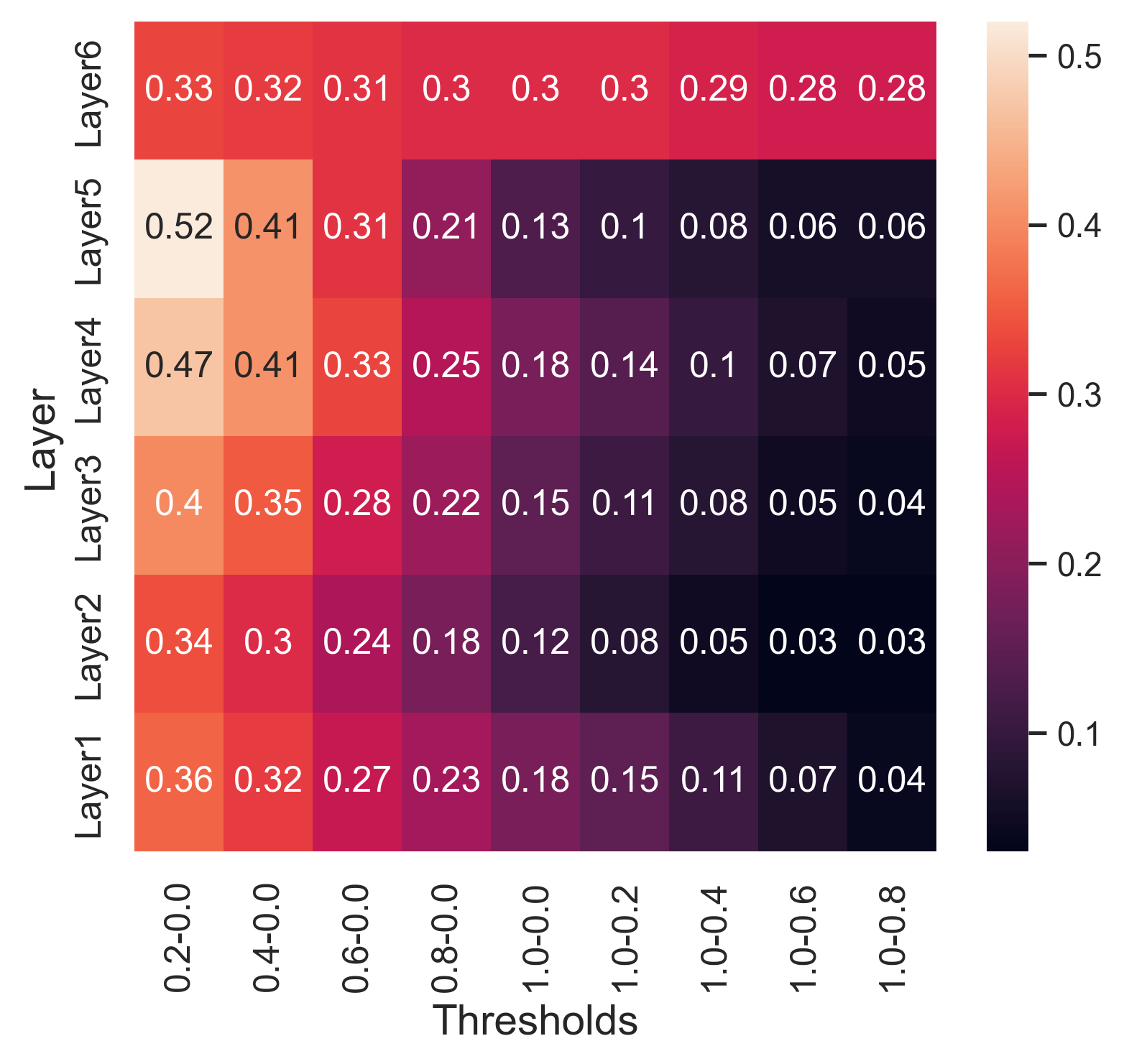}
  \caption{Confusion matrix of the average norm of the adapter representations in each layer of the decoder by the values of $\rho_{k_{min}}$ and $\rho_{k_{max}}$ on En-Vi.}
\label{fig_adapter_norm}
\end{figure}
\begin{figure}[H]
  \centering
\includegraphics[scale=0.45]{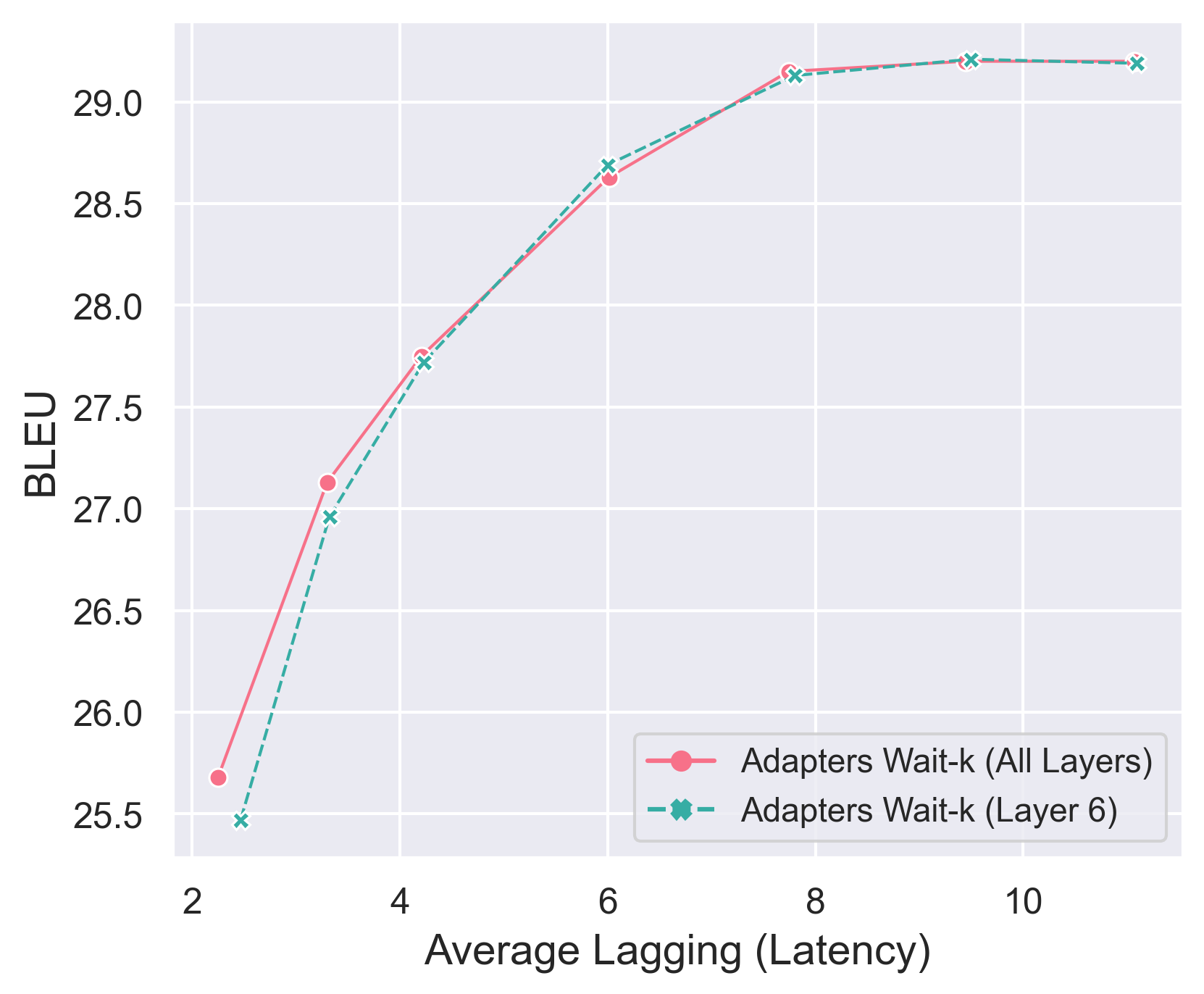}
  \caption{Comparison of the results of inserting adapters in all layers vs. only the last layer on En-Vi. We witness a drop in performance only in low latency levels.}
\label{fig_adapter_last_layer}
\end{figure}

We compute the L2-norm of the adapter representations in order to discover which adapter layers are involved in the representations \cite{liu-etal-2020-norm, zhu-etal-2021-counter-interference}. We measure the L2-norm during inference for $k_{min}=1$ and $k_{max}=9$ while varying the value of $\rho_{k_{min}}$ and $\rho_{k_{max}}$, as described in Section \ref{sec5.2}. As depicted in Figure \ref{fig_adapter_norm}, the norm for all layers except layer 6 decreases as we increase $\rho_{k_{min}}$ or $\rho_{k_{max}}$, which correlates with making the adaptive strategy more conservative because the threshold for making a write action is higher. This shows that the adapters are more involved in the prediction when the model is forced to be more aggressive. Only layer 6 is stably invested in adapting the model representations at all the threshold values, which seems to indicate that only low threshold predictions are complex enough to recruit all the adapter layers. Based on this observation, we experiment with inserting adapters only in the last layer (i.e. layer 6). We show in Figure \ref{fig_adapter_last_layer} the results of comparing between inserting adapters in all layers and inserting the adapters only in the last layer, where we see a drop in performance only in lower latency levels. This shows that we can make the model more efficient by removing lower layer adapters with a small drop in performance. 

\section{Conclusion}

In this paper, we employ adapters to build a SiMT model that can support multiple latency levels at inference. We use the multi-path training and show that by adding wait-k adapters we can flexibly balance parameter sharing and interference between the wait-k paths. Furthermore, we adopt a simple adaptive strategy and show that it further improves the results. By comparing against strong adaptive and fixed strategies, we find that our method achieves better or competitive results on most latency levels.

\section{Limitations}
The two datasets we used are common in SiMT research and were selected to compare against other baselines, but evaluating on only two language directions can be a limiting factor for the generalization of our results. Although Vietnamese is from a different language family, it deploys a similar word order (i.e. Subject-Verb-Object) to English and German and we believe that more challenges might emerge when dealing with language directions with a different word order. Additionally, we evaluate latency using common SiMT latency metrics such as AL, which are sentence-level and do not reflect the nature of a streaming scenario \cite{iranzo2021-stream}. Furthermore, in this work, we only evaluated on offline data, while evaluating on real interpretation data might offer more realistic results \cite{zhao2021-interpretation}.

\section*{Acknowledgements}
The research presented in this paper was conducted as part of VOXReality project\footnote{https://voxreality.eu/}, which was funded by the European Union Horizon Europe program under grant agreement No. 101070521.

\bibliography{acl2023}
\bibliographystyle{acl_natbib}

\appendix

\appendix
\label{sec:appendix}

\section{Hyperparameters}
We list the hyperparameters of our experiments in Table \ref{tab-hyperparameters}.
\begin{table*}[t]
\centering
\label{tab:hyperparameters}
\begin{tabular}{lccc}
\toprule
\textbf{Hyperparameter} & \textbf{IWSLT15 En$\rightarrow$Vi} & \textbf{WMT15 De$\rightarrow$En (Base)} & \textbf{WMT15 De$\rightarrow$En (Big)} \\
\midrule
Encoder layers & 6 & 6 & 6 \\
Encoder attention heads & 4 & 8 & 16 \\
Encoder embed dim & 512 & 512 & 1024 \\
Encoder FFN embed dim & 1024 & 2048 & 4096 \\
Decoder layers & 6 & 6 & 6 \\
Decoder attention heads & 4 & 8 & 16 \\
Decoder embed dim & 512 & 512 & 1024 \\
Decoder FFN embed dim & 1024 & 2048 & 4096 \\
Dropout & 0.3 & 0.3 & 0.3 \\
Optimizer & Adam & Adam & Adam \\
Adam-$\beta$ & (0.9, 0.98) & (0.9, 0.98) & (0.9, 0.98) \\
Clip-norm & 0. & 0. & 0. \\
Learning rate (lr) & 5e-4 & 5e-4 & 5e-4 \\
LR scheduler & inverse sqrt & inverse sqrt & inverse sqrt \\
Warm-up updates & 4000 & 4000 & 4000 \\
Warm-up init LR & 1e-7 & 1e-7 & 1e-7 \\
Weight decay & 1e-4 & 1e-4 & 1e-4 \\
Label smoothing & 0.1 & 0.1 & 0.1 \\
Max tokens & 16000 & 8192$\times$4 & 4096$\times$4$\times$2 \\
\bottomrule
\end{tabular}
\caption{System Hyperparameters}
\label{tab-hyperparameters}
\end{table*}

\section{Numeric Results}

In Tables \ref{table_envi}, \ref{table_deen_base} and \ref{table_deen_big}, we report the numeric results of our methods. We report the BLEU score for quality, while for latency we used Average Lagging (AL), Consecutive Wait (CW) \cite{gu-etal-2017-learning}, Average Proportion (AP) \cite{ChoE16} and Differentiable Average Lagging (DAL) \cite{arivazhagan-etal-2019-monotonic}. Below we provide the definition of CW, AP and DAL. $g(i)$ constitutes the number of tokens read when predicting $y_i$, while $|x|$ and $|y|$ refer to the number of source and target tokens respectively.

\textbf{Consecutive Wait (CW)} 
Computes the average number of consecutive tokens read between two predicted tokens.

\begin{equation}
CW =  \frac{\sum_{i=1}^{|y|} (g(i) - g(i - 1))}{\sum_{i=1}^{|y|} \mathbb{I}_{g(i) - g(i - 1) > 0}} 
\end{equation}

\textbf{Average Proportion (AP)}
Computes the proportion of tokens read to make every prediction.
\begin{equation}
    AP = \frac{1}{|x||y|} \sum_{i=1}^{|y|} g(i)
\end{equation}

\textbf{Differentiable Average Lagging (DAL)}
Is a differentiable version of the Average Lagging metric.
\begin{equation}
\begin{aligned}
    g'(i) = 
    \begin{cases}
        g(i) & \text{if } i = 1 \\ 
        \max\left(g(i), g'(i-1) + \frac{|x|}{|y|}\right) & \text{if } i > 1
    \end{cases}
\end{aligned}
\end{equation}

\begin{equation}
    DAL = \frac{1}{|y|} \sum_{i=1}^{|y|} g'(i) - \frac{i-1}{|x|/|y|}
\end{equation}

\begin{table*}[t]
  \centering
  \begin{tabular}{l|cccccc}
    \toprule
      \multicolumn{5}{c}{\textbf{IWSLT15 English→Vietnamese} \textbf{Transformer-Small}} \\
    \hline
     & \textbf{K} & \textbf{CW} & \textbf{AP} & \textbf{DAL} & \textbf{AL}  & \textbf{BLEU} \\    \multirow{6}{*}{\textbf{Adapters Wait-$k$}} & 1 & 1.16 & 0.59 & 3.32 & 2.25 & 25.68 \\
    & 2 &  1.17 & 0.64 & 4.13 & 3.30 &  27.13 \\
    & 3 & 1.22 & 0.68 & 4.91 & 4.21 & 27.75 \\
    & 5 & 1.44 & 0.75 & 6.63 & 6.01 &  28.63 \\
    & 7 & 1.87 & 0.81 & 8.36 & 7.74 & 29.15 \\
    & 9 & 2.56 & 0.85 & 10.05 & 9.45 &  29.20 \\
    \hline
    & \textbf{($\rho_{k_{min}}$, $\rho_{k_{max}}$)} & {\textbf{CW}} & {\textbf{AP}} & {\textbf{DAL}} & {\textbf{AL}}  & {\textbf{BLEU}} \\
    \multirow{9}{*}{\textbf{Adaptive Adapters}} & (0.2, 0.0) & 1.37 & 0.60 & 3.89 & 2.52 & 26.12 \\
    & (0.4, 0.0) & 1.73 & 0.63 & 5.04 & 3.13  & 27.24 \\
    & (0.6, 0.0) & 2.19 & 0.67 & 6.14 & 3.92 & 28.09 \\
    & (0.8, 0.0) & 2.66 & 0.71 & 6.95 & 4.80  & 28.62 \\
    & (1.0, 0.0) & 2.71 & 0.74 & 7.58 & 5.65 & 29.00 \\
    & (1.0, 0.2) & 3.08 & 0.76 & 8.40 & 6.36 & 29.08 \\
    & (1.0, 0.4) & 3.33 & 0.79 & 9.10 & 7.20  & 29.10 \\
    & (1.0, 0.6) & 3.34 & 0.82 & 9.55 & 8.01  & 29.18 \\
    & (1.0, 0.8) & 3.11 & 0.84 & 9.87 & 8.78  & 29.19 \\
    \bottomrule
  \end{tabular}

  \caption{Numerical results for En-Vi with Transformer-Small.}
  \label{table_envi}
\end{table*}

\begin{table*}[t]
\centering
\begin{tabular}{l|cccccc}
\toprule
      \multicolumn{5}{c}{\textbf{WMT15 German→English Transformer-Base}} \\
\hline
& \textbf{K} & \textbf{CW} & \textbf{AP} & \textbf{DAL} & \textbf{AL} & \textbf{BLEU} \\
\multirow{9}{*}{\textbf{Adapters \break Wait-$k$}} & 1 & 1.15 & 0.52 & 1.79 & 0.36 & 20.72 \\
& 2 & 1.19 & 0.55 & 2.49 & 1.00 & 23.37 \\
& 3 & 1.21 & 0.59 & 3.32 & 2.03 & 25.73 \\
& 5 & 1.37 & 0.66 & 5.19 & 3.85 & 27.71 \\
& 7 & 1.69 & 0.73 & 7.11 & 5.86 & 29.17 \\
& 9 & 2.16 & 0.78 & 8.98 & 7.76 & 30.05 \\
& 11 & 2.77 & 0.82 & 10.78 & 9.65 & 30.45 \\
& 13 & 3.52 & 0.85 & 12.49 & 11.46 & 30.90 \\
& 15 & 4.43 & 0.88 & 14.10 & 13.17 & 31.01 \\
\hline
& \textbf{($\rho_{k_{min}}$, $\rho_{k_{max}}$)} & {\textbf{CW}} & {\textbf{AP}} & {\textbf{DAL}} & {\textbf{AL}}  & {\textbf{BLEU}} \\
\multirow{9}{*}{\textbf{Adaptive Adapters}} & (0.2, 0.0) & 1.52 & 0.52 & 2.61 & 0.12 & 21.42 \\
& (0.4, 0.0) & 1.78 & 0.53 & 3.19 & 0.45 & 22.83 \\
& (0.6, 0.0) & 1.95 & 0.55 & 3.68 & 1.03 & 24.30 \\
& (0.8, 0.0) & 2.05 & 0.57 & 4.04 & 1.39 & 25.09 \\
& (1.0, 0.0) & 1.91 & 0.59 & 4.31 & 1.90 & 26.00 \\
& (1.0, 0.2) & 2.02 & 0.60 & 4.66 & 2.23 & 26.34 \\
& (1.0, 0.4) & 2.03 & 0.62 & 4.90 & 2.60 & 26.89 \\
& (1.0, 0.6) & 1.94 & 0.63 & 5.06 & 3.03 & 27.41 \\
& (1.0, 0.8) & 1.74 & 0.65 & 5.16 & 3.41 & 27.62 \\
\bottomrule
\end{tabular}
\caption{Numerical results for De-En with Transformer-Base.}
\label{table_deen_base}
\end{table*}

\begin{table*}[h]
\centering
\begin{tabular}{l|cccccc}
\toprule
      \multicolumn{5}{c}{\textbf{WMT15 German→English Transformer-Big}} \\
\hline
& \textbf{K} & \textbf{CW} & \textbf{AP} & \textbf{DAL} & \textbf{AL} & \textbf{BLEU} \\
\multirow{9}{*}{\textbf{Adapters Wait-$k$}} & 1 & 1.18 & 0.52 & 1.84 & 0.31 & 21.37 \\
& 2 & 1.19 & 0.55 & 2.55 & 1.09 & 24.53 \\
& 3 & 1.22 & 0.59 & 3.40 & 2.06 & 26.70 \\
& 5 & 1.38 & 0.66 & 5.24 & 3.88 & 28.98 \\
& 7 & 1.68 & 0.73 & 7.15 & 5.93 & 30.70 \\
& 9 & 2.16 & 0.78 & 9.02 & 7.85 & 31.50 \\
& 11 & 2.77 & 0.82 & 10.82 & 9.73 & 32.21 \\
& 13 & 3.52 & 0.85 & 12.52 & 11.50 & 32.31 \\
& 15 & 4.44 & 0.88 & 14.12 & 13.16 & 32.44 \\
\hline
& \textbf{($\rho_{k_{min}}$, $\rho_{k_{max}}$)} & {\textbf{CW}} & {\textbf{AP}} & {\textbf{DAL}} & {\textbf{AL}}  & {\textbf{BLEU}} \\
\multirow{9}{*}{\textbf{Adaptive Adapters}} & (0.2, 0.0) & 1.50 & 0.52 & 2.56 & 0.18 & 22.30 \\
& (0.4, 0.0) & 1.78 & 0.53 & 3.11 & 0.44 & 23.30 \\
& (0.6, 0.0) & 1.99 & 0.55 & 3.60 & 0.79 & 24.79 \\
& (0.8, 0.0) & 2.08 & 0.57 & 4.02 & 1.31 & 26.18 \\
& (1.0, 0.0) & 1.94 & 0.59 & 4.29 & 1.82 & 27.05 \\
& (1.0, 0.2) & 2.03 & 0.60 & 4.66 & 2.22 & 27.60 \\
& (1.0, 0.4) & 2.06 & 0.62 & 4.92 & 2.58 & 28.05 \\
& (1.0, 0.6) & 1.99 & 0.63 & 5.09 & 2.94 & 28.52 \\
& (1.0, 0.8) & 1.77 & 0.65 & 5.20 & 3.41 & 28.85 \\
\bottomrule
\end{tabular}
\caption{Numerical results for De-En with Transformer-Big.}
\label{table_deen_big}
\end{table*}

\end{document}